\documentclass[runningheads]{llncs}
\usepackage[T1]{fontenc}
\usepackage{amsmath}
\usepackage{amssymb}
\usepackage{graphicx}
\usepackage{float}
\usepackage[dvipsnames]{xcolor}
\usepackage{tcolorbox}
\usepackage{svg}
\usepackage{colortbl}
\usepackage{longtable}
\usepackage{array}
\usepackage{color}
\usepackage{tabularray}
\usepackage{breakcites}
\usepackage{hyperref}
\usepackage{lipsum}
\usepackage{tikz}
\usepackage{subcaption}
\usepackage{tcolorbox}
\usepackage{tabularray}
\begin{document}
\title{Seam Carving as Feature Pooling in CNN}
\titlerunning{Seam Carving in CNN}
%
\author{Mohammad Imrul Jubair}
%
\institute{
Computer Science, University of Colorado Boulder\\
\email{mohammad.jubair@colorado.edu}
}
\maketitle              

\begin{abstract}
This work investigates the potential of seam carving as a feature pooling technique within Convolutional Neural Networks (CNNs) for image classification, replacing the traditional max pooling layer with a seam carving operation. We train and evaluate both variants of a shallow CNN on a two-class bird classification task from the Caltech-UCSD Birds 200-2011 dataset, taking care that the two models being compared receive identical trainable capacity: since seam carving's seam-selection step is not end-to-end differentiable, we freeze the first convolutional layer identically in both models -- verified at runtime -- so that the comparison isolates the pooling operator itself rather than an incidental difference in what each model is able to learn. We also reimplement seam carving pooling as a batched, vectorized PyTorch operation, substantially reducing training time relative to a naive per-image implementation. Under this methodology, max pooling outperforms seam carving pooling on this task, achieving higher accuracy and a lower evaluation loss. We report accuracy, precision, recall, F1-score, and training cost for both models, and discuss why a content-aware pooling advantage, if one exists, is unlikely to be universal.

\end{abstract}

\noindent\textit{Note: an earlier version of this manuscript reported the opposite result, based on an implementation in which the two compared models did not have equal trainable capacity (Section~\ref{sec:fair}). This version corrects that issue and reports the resulting revised methodology and results throughout.}

\section{Introduction}
Feature pooling in Convolutional Neural Networks (CNNs) is a crucial process that involves summarizing or aggregating features from different spatial locations within a feature map. This operation is fundamental for reducing the dimensionality of the data, improving computational efficiency, and enhancing the network's ability to generalize \cite{PoolingTheir}.

\subsection{Max pooling}
Several pooling techniques have been developed to optimize feature extraction and improve network performance.
Max pooling is the most common and widely used pooling technique.
It first divides an input image or array into rectangular regions and selects the maximum value from each region.
It also introduces a degree of translation invariance, meaning the network is less sensitive to small shifts in the input image \cite{Zhao2024, Brownlee_2019}.

The idea of Max pooling approach has been extended in various ways.
For instance, Global max pooling performs pooling across the entire feature map, producing a single value per channel, which is particularly useful in the final layers of classification networks \cite{christlein2019deep}.
Fractional Maxpooling allows for downsampling by non-integer factors, offering greater flexibility in network design \cite{graham2014fractional}.
To introduce randomness and improve generalization, Stochastic Pooling selects values based on a probability distribution rather than always choosing the maximum~\cite{7301267}. Dilated Maxpooling expands the pooling window's reach by skipping elements, enabling the capture of more global context without losing spatial resolution, which is beneficial in tasks such as segmentation~\cite{yu2015multi}. Adaptive Maxpooling adjusts the pooling operation to produce a consistent output size regardless of input dimensions, making it ideal for networks processing variable-sized inputs~\cite{pytorchAdaptiveMaxPool2d}. Furthermore, Spatial Pyramid Pooling (SPP) divides the input into regions of varying sizes, pooling each and then concatenating the results, thereby preserving spatial information at multiple scales and proving useful in object detection~\cite{he2015spatial}. Multi-Scale Maxpooling combines pooling outputs from different scales to capture both fine and coarse features, enhancing the network's ability to process complex data like those required in semantic segmentation tasks~\cite{Yoo_2015_CVPR_Workshops}. These variations demonstrate the adaptability and effectiveness of Maxpooling in optimizing CNN performance across a wide range of applications.

\subsection{Seam Carving}
\begin{figure}[htb]
    \begin{subfigure}[h]{\linewidth}
        \centering
        \includegraphics[width=0.4\linewidth]{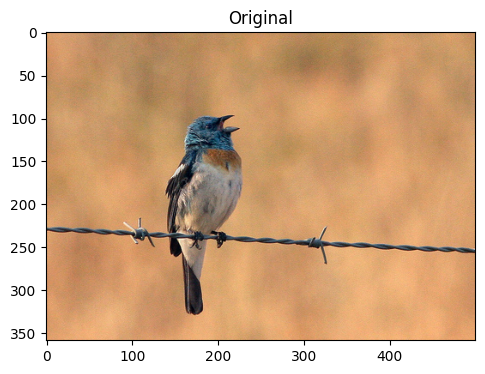}
        \label{fig:orig_img}
    \end{subfigure}
    \begin{subfigure}[h]{\linewidth}
        \centering
        \includegraphics[width=0.6\linewidth]{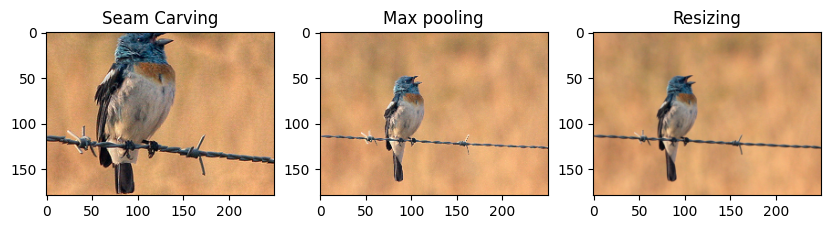}
        \label{fig:seam_max_resize}
    \end{subfigure}

    \caption{Example of Seam Carving, Max Pooling and Resizing (downsampling).}
    \label{fig:seam_vs_max_vs_resize}
\end{figure}

In this work, we highlight on Seam Carving algorithm~\cite{10.1145/1275808.1276390}.
Seam carving is an advanced content-aware image resizing technique that adapts an image's dimensions while preserving its most important features. Proposed by Avidan and Shamir in 2007, the method identifies and removes or inserts low-energy seams—paths of pixels that traverse the image from one edge to the opposite side with minimal impact on the overall visual content. This approach allows for more flexible image resizing, avoiding the distortion of critical regions that typically occurs with standard scaling methods. Seam carving is particularly useful for tasks like image retargeting, where maintaining the visual integrity of key elements, such as houses, cars or objects, is crucial~\cite{10.1145/1275808.1276390, kiess2010seam}. Figure~\ref{fig:seam_vs_max_vs_resize} provides examples to differentiate the output of downsampling using seam carving, max pooling and regular downsampling via resizing technique~\footnote{https://pytorch.org/vision/stable/generated/torchvision.transforms.Resize.html}.\\

\begin{figure}[htb]
    \centering
    \includegraphics[width=0.9\linewidth]{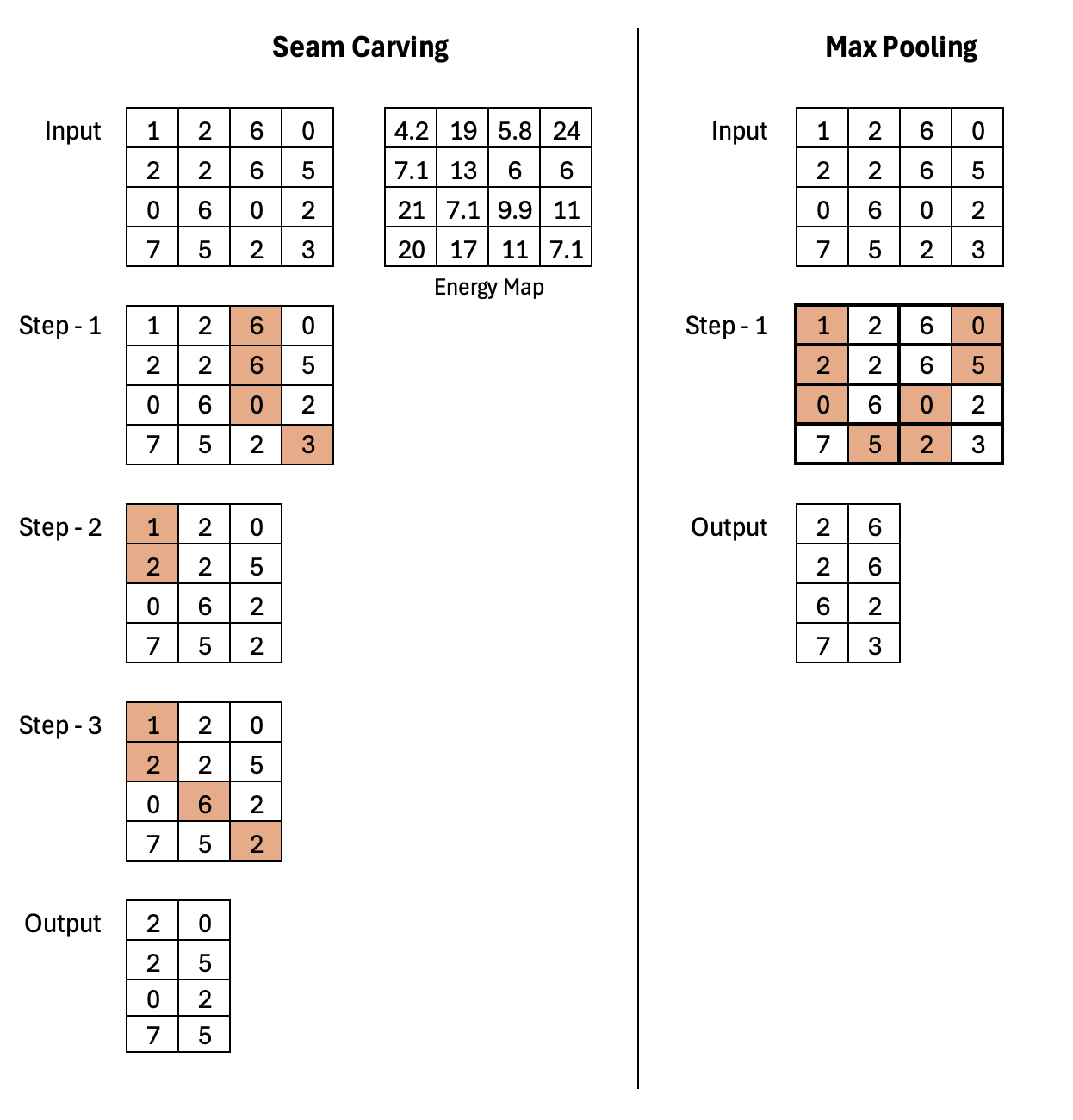}
    \caption{Examples illustration of the methodolgies of seam carving and max pooling. For seam carving, the energy maps is shown. And for max pooling, the corresponding windows are shown with bold border. For both, the orange items in the grids are selected to be removed during the processes.}

    \label{fig:seam_vs_max_procedure}
\end{figure}

In this work, we propose a feature pooling by leveraging seam carving technique. We simply replace the max pooling in CNN for image classification tasks.

\section{Methodology}

We first explain the fundamental workflow of seam carving. We also explain max pooling to differentiate betweem two algorithms.\\

Figure~\ref{fig:seam_vs_max_procedure} shows simplified illustration of the algorthims of seam carving and max pooling on a 2D array. In both the cases, input image is a $4\times 4$ matrix and our target is to reduce dimensions horizontally (i.e., removing columns). For seam carving, the energy map represents the importance or ``energy'' of each pixel. In this context, higher values (e.g., $24.2$) indicate more critical areas, while lower values (e.g., $4.2$) indicate less important areas that can be removed.
The algorithm then identifies a vertical seam (path) through the image where the sum of the energy values is the lowest. This seam is removed from the image (in the figure, the items in orange colors are removed from the matrix).
The process continues by removing multiple seams based on given number of iterations.
The resulting matrix after three seams have been removed is shown, which is a $4\times 2$ grid.

On the other hand, for max pooling technique, a $1\times 2$ pooling window moves across the input array, with the stride $(1, 2)$, the maximum value within that window is selected. The values which are in white are the maximums in their respective windows. The resulting output matrix after max pooling is a $4\times 2$ grid.

\subsection{Seam Carving integration in CNN}

\begin{figure}[htb]
    \centering
    \includegraphics[width=\linewidth]{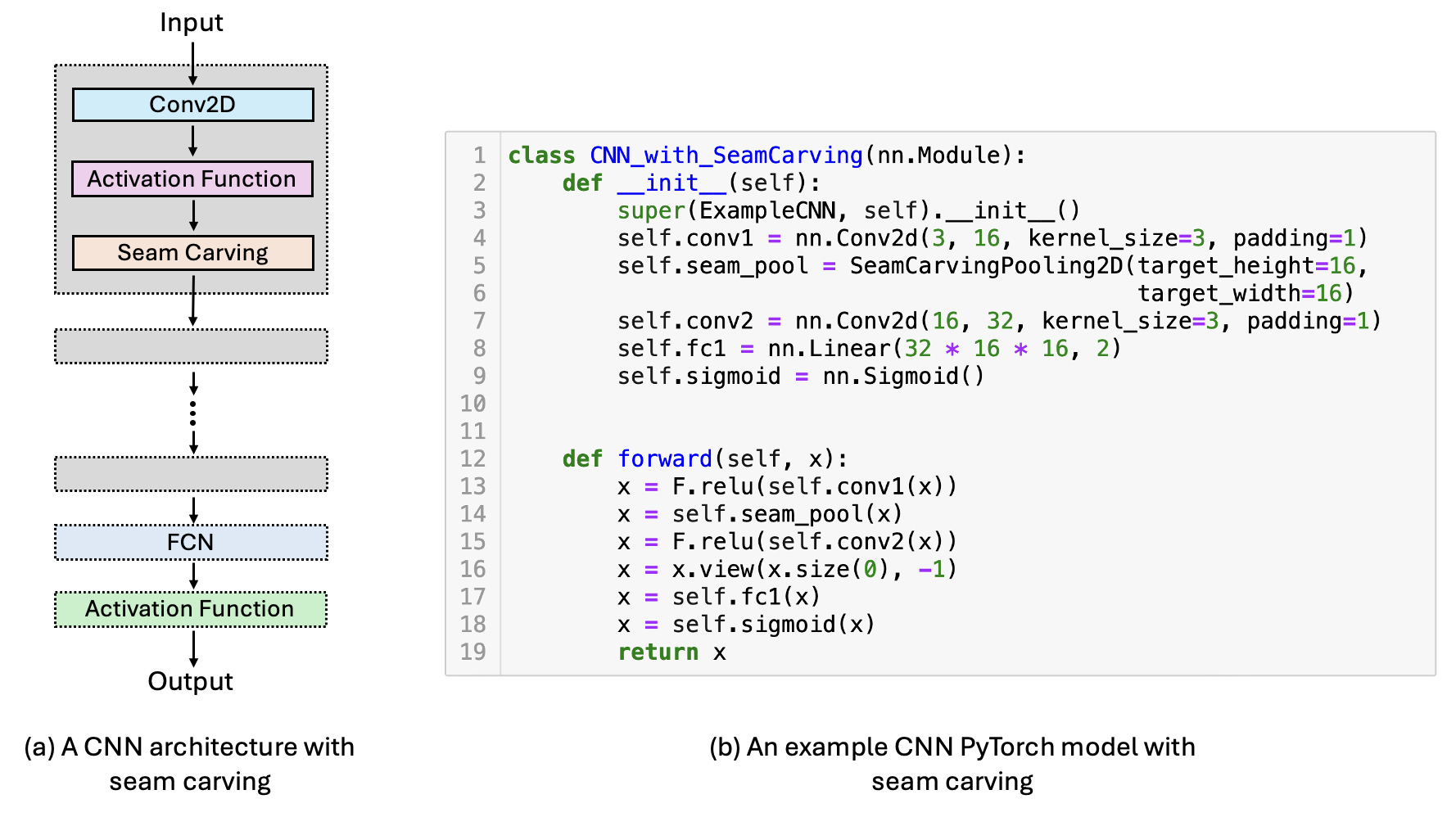}
    \caption{Integration of seam carving as feature pooling in CNN. (a) shows a basic CNN architecture blocks, indicating the use of seam carving layer. (b) shows an example CNN PyTorch model with seam carving function.}

    \label{fig:seam_architecture}
\end{figure}

In our modified CNN architecutre, we replace max pooling with seam carving. The architecutre first takes the image as input and 2D convolution is applied following seam carving. Figure~\ref{fig:seam_architecture}(a) presents the basic building block of a CNN architecutre with the integration of seam carving. We also presented a \texttt{PyTorch} implementation of the model. The main difference between our model and the traditional model is---we replaced max pooling with seam carving algorithm. Our implementation, including the fairness verification described in Section~\ref{sec:fair}, is available at \url{https://github.com/imruljubair/seam}.

\subsection{Ensuring a Fair Comparison}\label{sec:fair}

Seam carving's seam-selection step involves an $\arg\min$ over a dynamic-programming energy accumulation, which has no meaningful gradient: the seam carving pooling layer is not end-to-end differentiable, regardless of implementation. This has a direct consequence for training that is easy to overlook: if the seam-carving model's earlier layers (\texttt{conv1} in Figure~\ref{fig:seam_architecture}(b)) are trained via a standard backward pass, no gradient reaches them through the pooling layer, so those layers never update from their random initialization -- while the corresponding layer in a max-pooling model, which \emph{is} fully differentiable (max pooling routes the gradient to the position of the maximum value in each window), trains normally. Left unaddressed, this gives the two models being compared different trainable capacity -- max pooling with three fully trainable layers, versus seam carving with only two -- which confounds any comparison of the pooling operators themselves with a comparison of how much of the network each one lets you train.

To isolate the pooling operator as the only difference between the two models, we freeze \texttt{conv1} (weights and biases; \texttt{requires\_grad=False}) identically in \emph{both} models. Since both are initialized from the same random seed, this makes their frozen \texttt{conv1} weights bit-identical: both pooling operators see exactly the same low-level features, and the only remaining difference between the two models is the pooling operator itself. We verify this invariant at runtime for every training run reported in this paper: we snapshot \texttt{conv1}'s weights before training and assert they are numerically unchanged afterward, and, symmetrically, assert that \texttt{conv2}'s weights \emph{do} change, confirming gradients flow to the rest of the network in both models. We recommend this kind of runtime check as standard practice whenever a paper introduces a custom layer and compares it against a fully-trainable baseline, since a layer that silently blocks gradient flow does not raise an error or a warning during training -- loss decreases either way, and nothing in standard training output reveals that part of a network has stopped learning.

We additionally use raw logits with \texttt{CrossEntropyLoss} (which applies \texttt{log\_softmax} internally) rather than applying a \texttt{Sigmoid} activation beforehand, which would otherwise redundantly saturate gradients before the loss.

\subsection{Vectorized Implementation}\label{sec:vectorized}

We implement the seam-carving pooling layer as a batched PyTorch operation rather than looping over each image in the batch with per-image NumPy/SciPy calls. The energy function is computed for the whole batch with a single Sobel convolution; the dynamic-programming seam search is vectorized across the batch and across spatial columns, leaving only the (inherently sequential) sweep over image rows as a Python-level loop; and seam removal is implemented via a boolean mask and reshape rather than per-row Python loops. We verified numerical agreement with a reference per-image implementation on randomized inputs. This substantially reduces wall-clock training time relative to a naive implementation (Section~\ref{sec:cost}).

\section{Experiments and Results}

\subsection{Dataset and Task}
In this article, we experimented seam carving based CNN model on the \textit{Caltech UCSD Birds 200-2011} dataset~\cite{wah2011caltech}.
The dataset contains $11788$ images of $200$ bird species. Each species is associated with a Wikipedia article and organized by scientific classification.

We aim to perform classification task to evaluated our model. From the classes in the dataset, we considered only two categories of birds, which are \textit{Bobolink} (Class - 0) and \textit{Indigo Bunting} (Class - 1), 60 images per class. RGB images are resized to $32\times32$ and converted to tensors. We use an $80{:}20$ train/validation split, and report accuracy, precision, recall and F1-score on the validation split (24 images: 13 Bobolink, 11 Indigo Bunting).

\begin{figure}[htb]
    \centering
    \includegraphics[width=.9\linewidth]{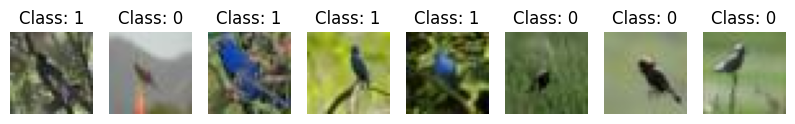}
    \caption{Few examples from the classes: \textit{Bobolink} (Class - 0) and \textit{Indigo Bunting} (Class - 1).}
    \label{fig:classes}
\end{figure}

\subsection{Architecture}
We designed a template architecture and make a two version of it: one with max pooling and another with seam carving. In this work, we used the architecture presented in Figure \ref{fig:seam_architecture}(b). In case of max pooling, we simply used \texttt{nn.MaxPool2d (kernel\_size = (2, 2) )} in \texttt{line 5}.

\begin{figure}[htb]
    \centering
    \includegraphics[width=0.93\linewidth]{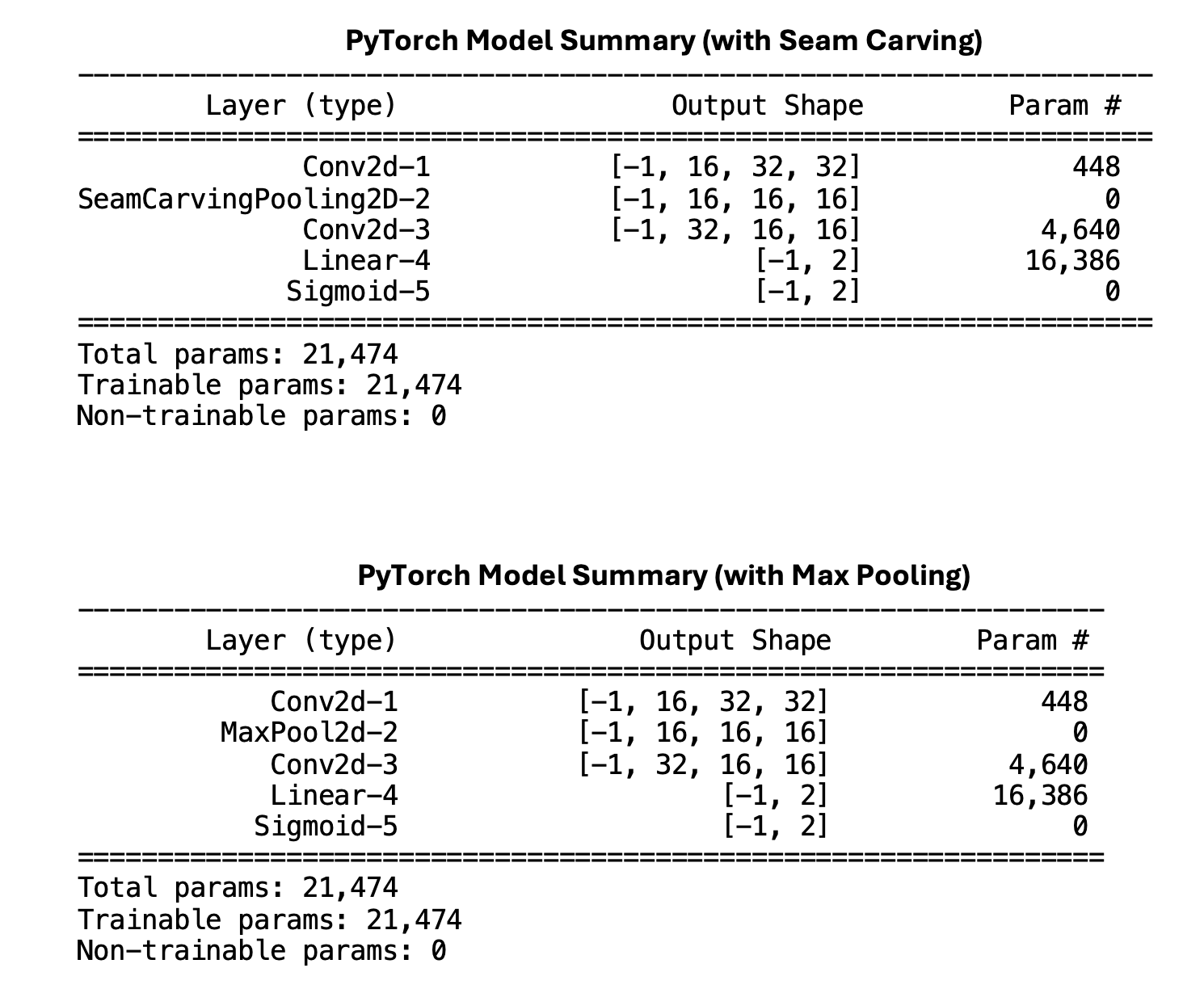}
    \caption{\texttt{PyTorch} model summary.}
    \label{fig:torchsummary}
\end{figure}

We conducted experiments utilizing a relatively shallow model. The model summary can be found in Figure~\ref{fig:torchsummary}. The output of the first convolutional layer has 16 channels and each channel is of size $32\times32$. After the pooling layer (Max pooling or Seam Carving), the output has $16$ channels with each channel being reduced to a size of $16\times16$. The output of the second convolutional layer has $32$ channels, with each channel being $16\times16$. We used \textit{ReLU} function after each convolution operation, and the output is fed into a fully connected layer producing raw class logits (Section~\ref{sec:fair}). \texttt{conv1} is frozen identically in both models as described in Section~\ref{sec:fair}. We used \textit{Cross Entropy} as loss function and \textit{Stochastic Gradient Descent (SGD)} as optimizer.

\subsection{Training and Evaluation Metric}

We applied early stopping mechanism based on validation performance.
To ensure consistency and comparability, we utilized the same random seed for initializing the weights of both models (one with seam carving and the other with max pooling); as described in Section~\ref{sec:fair}, this also guarantees bit-identical frozen \texttt{conv1} weights between the two models, which we assert at runtime. Table~\ref{tab:hyp} shows the setup for other hyper-parameters. Accuracy, precision, recall and F1 score are used for measuring the models' performance.

\begin{table}[htb]
    \centering
    \caption{Hyper Parameters}
    \label{tab:hyp}
    \small
    \begin{tblr}{
      width = \linewidth,
      colspec = {Q[130]Q[100]},
      column{1} = {r},
      hline{1,2,7} = {-}{}
    }
    \textbf{Parameters} & \textbf{Values} \\
    \texttt{Random Seed}                & \texttt{12}              \\
    \texttt{Learning Rate}      &  \texttt{0.01}            \\
    \texttt{Traininng Batch Size}         &  \texttt{16}              \\
    \texttt{Maximum Epochs}    &  \texttt{300}             \\
    \texttt{Early Stopping Patience}            &  \texttt{25}
    \end{tblr}
\end{table}

\subsection{Results}

\begin{figure}[htb]
    \centering
    \includegraphics[width=\linewidth]{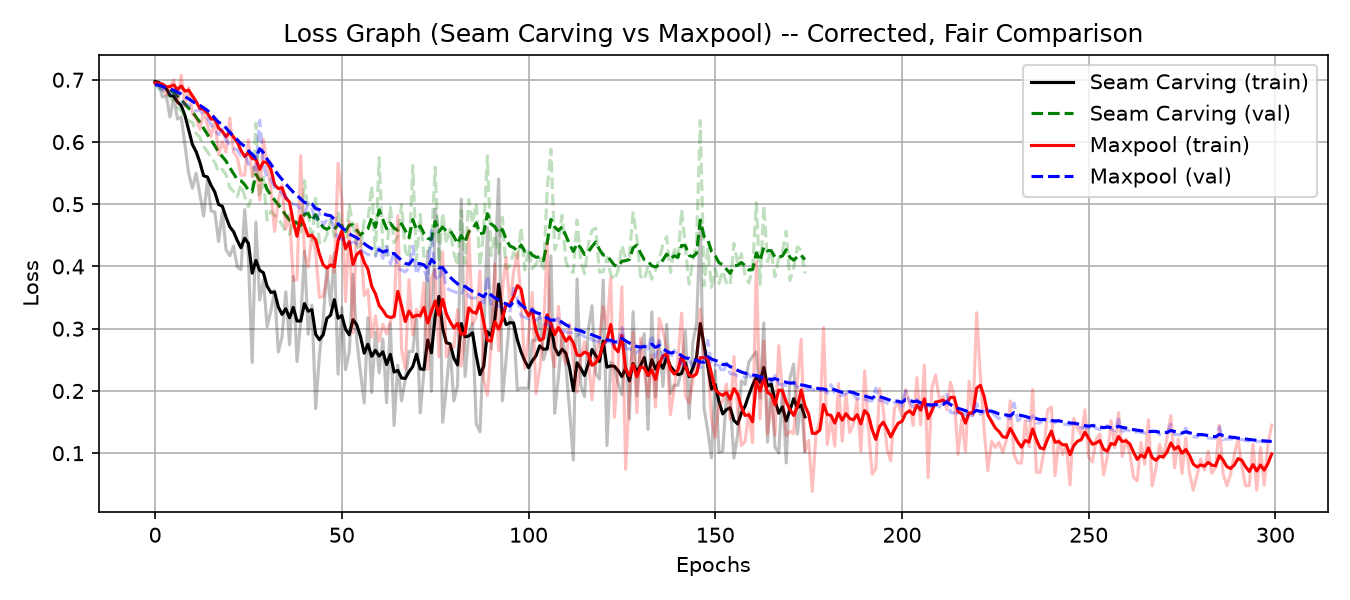}
    \caption{Loss graph of CNN models with max pooling and with seam carving.}

    \label{fig:lossgraph}
\end{figure}

Figure~\ref{fig:lossgraph} illustrates the loss graph for training both models. Seam carving's training loss falls faster and lower than max pooling's early on, but its validation loss plateaus around epoch 50 with persistent noisy fluctuation and does not meaningfully improve afterward, a pattern consistent with overfitting on top of a fixed, frozen feature extractor. Max pooling's validation loss, in contrast, decreases steadily and smoothly for the full 300 epochs, ending substantially lower. Seam carving reaches its best evaluation loss of $0.366$ at epoch 149 before early stopping at epoch 174; max pooling reaches its best evaluation loss of $0.117$ at its final epoch (300), still improving when training ended.

\begin{figure}[htb]
    \centering
    \includegraphics[width=0.9\linewidth]{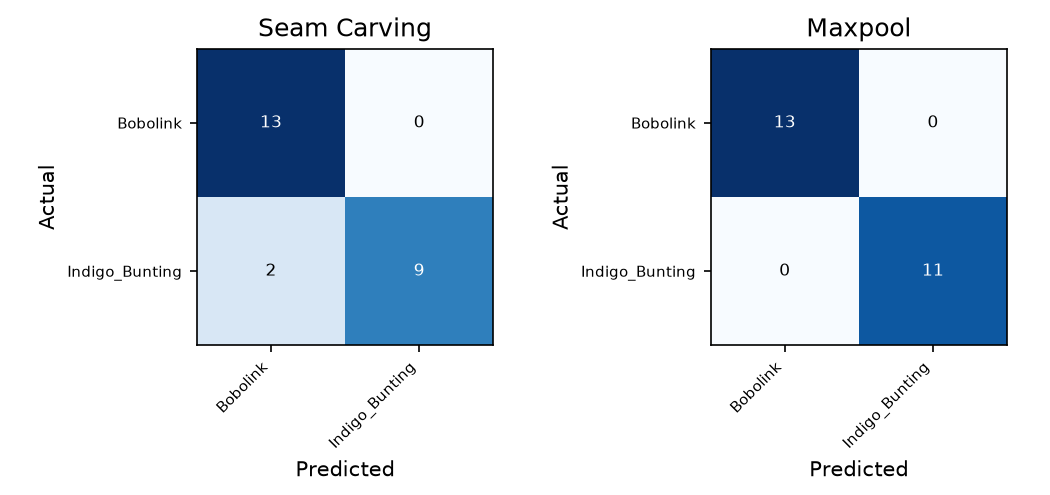}
    \caption{Confusion matrix. For the CNN models with seam carving (left), and with max pooling (right).}

    \label{fig:cfmatrix}
\end{figure}

Figure~\ref{fig:cfmatrix} shows the confusion matrices for the bird classification task. Max pooling reaches \textbf{100\%} accuracy on the validation split; seam carving reaches \textbf{91.7\%}.
Table~\ref{tab:metric} gives the full precision/recall/F1 breakdown: max pooling is at or above seam carving on every metric for both classes.

\begin{table}[htb]
    \centering
    \small
    \caption{Performance Analysis.}
    \label{tab:metric}
    \begin{tblr}{
      width = \linewidth,
      colspec = {Q[137]Q[188]Q[208]Q[188]Q[208]},
      cells = {c},
      cell{1}{2} = {c=2}{0.396\linewidth},
      cell{1}{4} = {c=2}{0.396\linewidth},
      vline{2-3} = {1}{},
      vline{2,4} = {2-5}{},
      hline{2} = {2-5}{},
      hline{3} = {-}{},
    }
              & \textbf{\textit{Bobolink}} &                       & \textbf{\textit{Indigo Bunting}} &                       \\
              & \textbf{Max pooling}        & \textbf{Seam carving} & \textbf{Max pooling}        & \textbf{Seam carving} \\
    Precision & 1.00                        & 0.87                  & 1.00                        & 1.00                  \\
    Recall    & 1.00                        & 1.00                  & 1.00                        & 0.82                  \\
    f1-score  & 1.00                        & 0.93                  & 1.00                        & 0.90
    \end{tblr}
\end{table}

\subsection{Computational Cost}\label{sec:cost}

A naive, per-image NumPy/SciPy implementation of seam carving pooling (looping over each image in the batch) is markedly slower than max pooling: for the shallow, single-pooling-layer network used in this paper, such an implementation takes on the order of minutes per 100 epochs even on a GPU, versus under a second for max pooling. Our vectorized reimplementation (Section~\ref{sec:vectorized}), which processes an entire batch's seam removal at once, reduces this substantially: on this paper's hyperparameters (batch size 16, 6 training batches/epoch), a full training epoch with the vectorized seam-carving model takes approximately 0.5 seconds on a consumer CPU (Apple M2, no GPU), roughly two orders of magnitude faster than the naive per-image approach's epoch cost, on weaker hardware.

On the larger, full 200-class CUB-200-2011 dataset (11{,}788 images, batch size 16, $\sim$590 training batches/epoch), we measured 61 seconds/epoch for seam carving and 23 seconds/epoch for max pooling on the same CPU. Profiling revealed that for max pooling, this time was dominated not by model computation (under 4 ms/batch) but by single-threaded JPEG decoding in the data loader. Enabling parallel data loading (4 worker processes, \texttt{persistent\_workers=True} to avoid macOS's per-epoch worker-respawn cost) reduced max pooling's epoch time to 7.2 seconds ($3.2\times$) and seam carving's to 45.8 seconds ($1.3\times$); seam carving benefits less because it remains dominated by its own sequential per-seam computation rather than data loading.

We also benchmarked Apple's Metal Performance Shaders (MPS) GPU backend, available on Apple Silicon, expecting it to accelerate seam carving pooling as it does standard convolutional layers. It does not: MPS was $3\times$ \emph{slower} than CPU for the seam-carving pooling layer alone (176ms vs.\ 56ms per batch of 16 images), and CPU remained faster even at batch sizes up to 1024. We attribute this to the layer's sequential structure -- roughly $H+W$ Python-level iterations per forward pass, each performing several small tensor operations -- which pays MPS's per-operation dispatch overhead without enough parallel work per operation to amortize it. MPS did accelerate the max-pooling model ($2.3\times$), but since seam carving pooling remains the dominant cost when both models are considered together, forcing MPS for both would make total wall-clock time worse, not better; our released code defaults to CPU and exposes device selection as an explicit option.

\section{Limitations}

As noted by Garg et al. \cite{garg2023analysis}, seam carving techniques possess inherent limitations that inevitably impact the performance of CNN architectures incorporating them. This section presents the key challenges associated with using seam carving as a feature pooling method in CNNs.

Seam carving's computational complexity remains higher than traditional max pooling despite the vectorized reimplementation in Section~\ref{sec:vectorized}: it still requires an inherently sequential, per-seam dynamic-programming search, and (Section~\ref{sec:cost}) is 1.3--3$\times$ slower than max pooling per training epoch depending on how much of the total cost is data loading versus computation. Deeper networks, or networks with pooling at multiple stages, would further exacerbate this overhead.

The non-uniform downsampling nature of seam carving can distort important features and disrupt spatial relationships within the image. This distortion may negatively impact classification accuracy, especially in tasks that rely heavily on precise spatial information.

The effectiveness of seam carving is likely highly dataset-specific: it depends on there being a genuine foreground/background energy contrast for it to exploit. In this study, the predominance of natural backgrounds in the bird images gave seam carving relatively favorable conditions to identify insignificant seams, yet it still did not outperform max pooling. For images with complex scenes, multiple objects, or occluded/irregularly shaped subjects, seam carving may struggle even more to determine optimal seams without compromising image quality.

This paper's empirical validation remains narrow: a single dataset, two classes, a single shallow architecture, and a single random seed. We have not evaluated the full 200-class CUB-200-2011 task, or datasets with different visual characteristics (e.g., simpler, more uniform backgrounds, or more cluttered scenes). Further research across more datasets, architectures, and seeds is needed before drawing general conclusions about content-aware pooling.

The compatibility of seam carving with modern CNN techniques, such as batch normalization, dropout, and skip connections, requires thorough investigation to ensure seamless integration and optimal performance.

Finally, seam carving pooling as implemented here is not end-to-end differentiable: the seam-selection step involves a hard $\arg\min$ with no defined gradient (Section~\ref{sec:fair}). Freezing \texttt{conv1} in both models makes the comparison in this paper fair, but it does not make seam carving pooling differentiable, so these results should not be read as evidence about how seam carving pooling would perform if all layers, including those below it, could be trained end-to-end.

\section{Conclusion and Future Work}
This study investigates the potential of seam carving as a feature pooling technique within CNNs for image classification tasks. We designed CNN architectures incorporating seam carving and compared their performance to models utilizing traditional max pooling, taking care that both models receive identical trainable capacity (Section~\ref{sec:fair}) and reimplementing seam carving pooling as a vectorized, batched operation (Section~\ref{sec:vectorized}). The \textit{Caltech-UCSD Birds 200-2011} dataset was employed for bird classification, providing a benchmark for evaluating the efficacy of seam carving-based pooling.

Under this methodology, max pooling outperformed seam carving pooling: higher accuracy (100\% vs.\ 91.7\%) and a lower evaluation loss (0.117 vs.\ 0.366) on the validation split. We do not conclude that content-aware pooling can never be useful in a CNN; we conclude that this particular formulation of it, evaluated as carefully as we are currently able to, does not outperform max pooling on the task tested here. This research is limited in scope, and further investigation is necessary to comprehensively assess the generalizability of seam carving as a feature pooling technique.

Potential avenues for future research include:
\begin{itemize}
    \item \textit{Differentiable seam selection:} seam carving pooling is not end-to-end differentiable regardless of the \texttt{conv1}-freezing methodology in this paper (Section~\ref{sec:fair}); a soft/relaxed seam selection or a straight-through gradient estimator would let all layers train, and would test a different, arguably more interesting question than the one this paper answers.
    \item \textit{Dataset Diversity:} Evaluating the performance of these techniques on a broader range of datasets -- including the full 200-class CUB-200-2011 task, and datasets with different visual characteristics than natural bird photographs -- would provide valuable insights into their generalizability and applicability across different domains.
    \item \textit{Hybrid Approaches:} Combining seam carving and max pooling techniques might leverage the strengths of both methods, potentially leading to better results than either alone.
    \item \textit{Hyperparameter Optimization:} Exploring various hyperparameter settings for both seam carving and max pooling techniques could potentially enhance their performance.
\end{itemize}

\subsubsection*{Code and Data Availability} Our implementation -- including the runtime fairness verification described in Section~\ref{sec:fair} and the exact commands used to produce every result in this paper -- is available at \url{https://github.com/imruljubair/seam}.

\bibliographystyle{splncs04}
\bibliography{references}

%
\end{document}